\pdfoutput=1

\documentclass[11pt]{article}

\usepackage[]{ACL2023}

\usepackage{times}
\usepackage{latexsym}
\usepackage{amsmath}
\usepackage{multirow}
\usepackage{bm}
\usepackage{tabularx}
\usepackage{booktabs}
\usepackage{amsfonts}
\usepackage{algorithm}
\usepackage{algorithmic}
\usepackage{graphicx}
\usepackage{CJK}
\usepackage{subfigure}
 
\usepackage[T1]{fontenc}

\usepackage[utf8]{inputenc}

\usepackage{microtype}

\usepackage{inconsolata}

\title{QASnowball: An Iterative Bootstrapping Framework for \\ High-Quality Question-Answering Data Generation}

\author{
  Kunlun Zhu\(^{*1,2}\), Shihao Liang\(^{*1,2}\), Xu Han\(^{*1}\), Zhi Zheng\(^{2}\), Guoyang Zeng\(^{2}\) \\ \textbf{Zhiyuan Liu}\(^{\dagger1}\), \textbf{Maosong Sun}\(^{\dagger1}\) \\
  \textit{Tsinghua University\(^1\)}, \textit{ModelBest Inc.\(^{2}\)} \\
}

\begin{document}
\maketitle
\begin{abstract}

Recent years have witnessed the success of question answering (QA), especially its potential to be a foundation paradigm for tackling diverse NLP tasks. However, obtaining sufficient data to build an effective and stable QA system still remains an open problem. For this problem, we introduce an iterative bootstrapping framework for QA data augmentation (named QASnowball), which can iteratively generate large-scale high-quality QA data based on a seed set of supervised examples. Specifically, QASnowball consists of three modules, an answer extractor to extract core phrases in unlabeled documents as candidate answers, a question generator to generate questions based on documents and candidate answers, and a QA data filter to filter out high-quality QA data. Moreover, QASnowball can be self-enhanced by reseeding the seed set to fine-tune itself in different iterations, leading to continual improvements in the generation quality. We conduct experiments in the high-resource English scenario and the medium-resource Chinese scenario, and the experimental results show that the data generated by QASnowball can facilitate QA models: (1) training models on the generated data achieves comparable results to using supervised data, and (2) pre-training on the generated data and fine-tuning on supervised data can achieve better performance. Our code and generated data will be released to advance further work.


\end{abstract}

\section{Introduction}
\renewcommand{\thefootnote}{}

\footnotetext{\(^{*}\)Indicates equal contribution.}
\footnotetext{\(^{\dagger}\)Corresponding author.}

\renewcommand{\thefootnote}{\arabic{footnote}}
Recent years have witnessed the success of question answering (QA), such as machine reading comprehension (MRC)~\cite{liu2019neural,baradaran2022survey} and open-domain question answering (ODQA)~\cite{chen2017reading,karpukhin2020dense,zhu2021retrieving}. In addition to the development of QA itself, many efforts have been devoted to re-formalizing various NLP tasks into the QA form to better handle these tasks, such as dialogue systems~\cite{thoppilan2022lamda} and information extraction~\cite{li2020unified,liu2020event}. The great performance of the recently proposed GPT-3~\cite{brown2020language} and ChatGPT~\footnote{\url{https://openai.com/blog/chatgpt/}} further show the feasibility and effectiveness of using QA as the foundation to handle multiple diverse NLP tasks.

Despite the potential of QA to become a fundamental paradigm, manually obtaining sufficient data to train an effective and stable QA model is time-consuming and labor-intensive. To this end, we introduce an iterative bootstrapping framework to automatically generate QA data. 
Different from existing non-iterative methods that directly generate large-scale data at one time, without considering the continual improvement to data quality~\cite{alberti2019synthetic,lewis2021paq}, we build our framework on a supervised seed set and then iteratively generate data. During iterative generation, all generated data can be used to modify the seed set and enhance the entire framework, leading to continually generating data with higher quality and larger scale. \citet{agichtein2000snowball} refer to this process as snowball, and for convenience, we name our framework ``QASnowball''.

\begin{figure*}[t]
    \centering
    \includegraphics[width=1.0\linewidth]{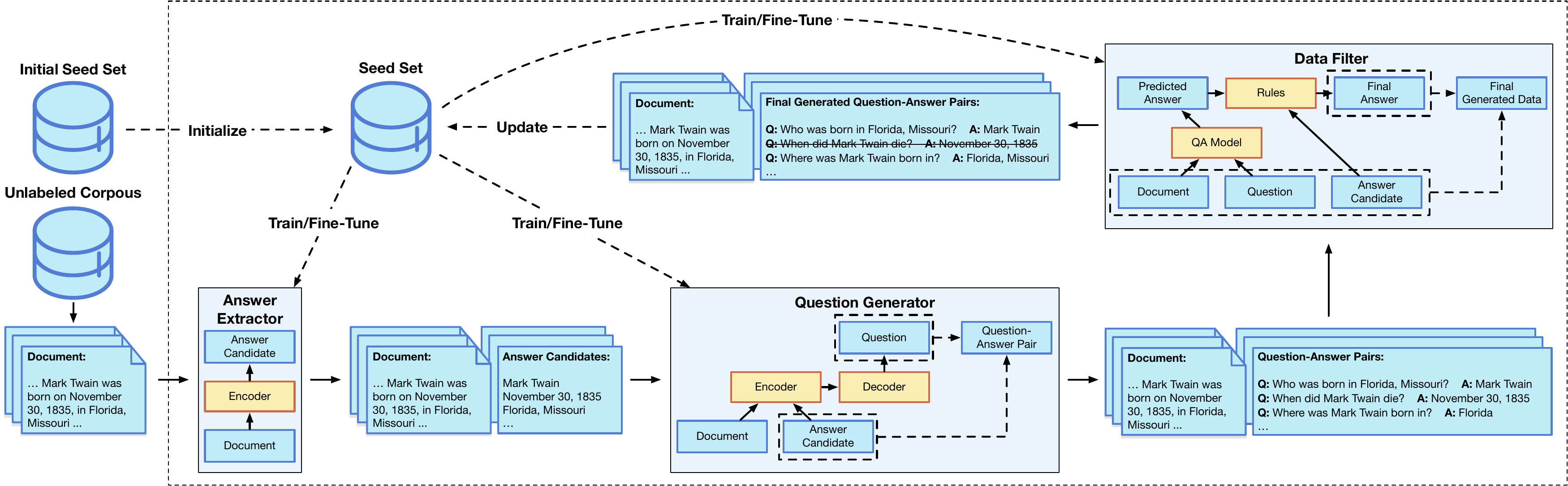}
    \caption{The overall framework of QASnowball.}
    \label{fig:model-large}
\end{figure*}

As shown in Figure~\ref{fig:model-large}, QASnowball divides the process of generating data into three steps and builds three corresponding modules: answer extractor, question generator, and data filter. All these modules are trained on the seed set. During data generation, given an unlabeled document, the extractor is first employed to extract the core phrases in the document that are most likely to work as answers. These extracted phrases serve as answer candidates for the subsequent question generation and data filtering. Then, the question generator is used to generate a corresponding question for each answer candidate, and the data filter is further applied to determine whether the generated questions match the documents and the answer candidates. The well-matched question-answer pairs are combined with the document as the final generated data.

Considering that the initial seed set includes only a portion of supervised QA samples, relying exclusively on the initial seed set may make the generated data lack diversity and coverage. To avoid this issue, QASnowball adopts an iterative bootstrapping method to make the framework self-enhanced. Given an unlabeled corpus, instead of directly generating QA data based on the corpus, QASnowball divides the corpus into several pieces and generates data based on only one piece at each iteration. After completing the generation process of each iteration, the generated data is integrated into the seed set to fine-tune the modules of QASnowball. Through this self-enhanced mechanism, QASnowball can continually improve the quality of the generated data and make the generated data more diverse than the initial seed set.

To evaluate the effectiveness of QASnowball, we conduct experiments in both the high-resource English and medium-resource Chinese scenarios. We choose the recent competitive QA model PERT~\cite{cui2022pert} and train PERT on different datasets,  including our auto-labeled data and some existing supervised and auto-labeled datasets. By comparing the results of PERT trained on different datasets, we can evaluate the impact of these datasets on the performance of QA models. The experimental results show that:
\begin{itemize}
\item Training PERT only on the auto-labeled data can achieve comparable results to training PERT on existing supervised datasets, indicating the high quality of the generated data. 
\item Pre-training PERT on the auto-labeled data and then fine-tuning the pre-trained PERT on the supervised data can outperform using only the supervised data, demonstrating the generated data can supplement the supervised data. 
\item Compared with the recently proposed auto-labeled data PAQ~\cite{lewis2021paq}, the data generated by QASnowball can help the QA model achieve better performance, indicating the effectiveness of the self-enhanced mechanism.
\end{itemize}
Some empirical analyses are further conducted to reveal the relation between the data quality and the iteration mechanism in a more fine-grained manner.
To advance the exploration of effective data augmentation methods, especially for the QA task, the code and generated data of QASnowball will be publicly available.

\section{Related Work}

\subsection{Question Answering}
Question Answering (QA) aims to answer the given questions in natural language. Based on the presence of contextual information, QA can be further categorized into two tasks, i.e. machine reading comprehension (MRC)~\citep{baradaran2022survey} and open-domain QA (ODQA)~\citep{zhu2021retrieving}. Specifically, MRC requires machines to answer questions based on the given document, while ODQA aims to answer factoid questions without any context. In this paper, we choose the MRC task to evaluate the effectiveness of the proposed method, due to its simplicity and interpretability.

\subsection{Question Generation and QA Data}

Question Generation (QG) aims to generate questions based on the given document and answers, which has been applied to various tasks, such as data augmentation~\citep{alberti2019synthetic} and document retrieval~\citep{nogueira2019document}. Some prior works also leverage question generation to generate QA data, such as Ocean-Q~\citep{fang2020accelerating} and PAQ~\citep{lewis2021paq}. 
However, these works focus on directly generating large-scale data at one time, regardless of the noise issue in the generated datasets. To this end, we introduce an iterative bootstrapping framework for high-quality QA data generation, which can alleviate the noise issue during data generation.

\subsection{Self-enhanced bootstrapping system}
Snowball is first introduced by \citet{agichtein2000snowball} for the relation extraction task, which propose a system to generate extraction patterns and new seed tuples in each iteration, and evaluate the quality of them automatically. Other similar bootstrapping systems have since been introduced, such as StatSnowball\citep{zhu2009statsnowball} and Neural Snowball\citep{gao2020neural}, both of which aim to improve relation extraction through iterations. To the best of our knowledge, we are the first to use a Snowball-like framework to generate QA data and demonstrate its effectiveness on many datasets.

\section{QASnowball}

In this section, we will introduce the details of our iterative bootstrapping framework QASnowball.

\subsection{The Overall Framework}

The overall framework of QASnowball consists of three modules: an answer extractor, a question generator, and a data filter. As shown in Figure~\ref{fig:model-large}, these modules work in a pipeline to automatically generate QA data.

Specifically, given an unlabeled document $D$: 
(1) \textbf{The answer extractor} first extracts the core phrases $\mathcal{A}_D = \{A_D^1, A_D^2, \ldots, A_D^n\}$ that are most likely to be answered in $D$ as answer candidates. 
(2) \textbf{The question generator} generates the question set $\mathcal{Q}_D = \{Q_D^1, Q_D^2, \ldots, Q_D^n\}$ corresponding to the answer set $\mathcal{A}_D$, according to both the semantics of the given document and the answer candidates, where $Q_D^i$ is corresponding to $A_D^i$. 
(3) \textbf{The data filter} selects high-quality question-answer pairs from the combinations of $\mathcal{A}_D$ and $\mathcal{Q}_D$, and the selected question-answer pairs are combined with $D$ to form the final generated data.

All modules of QASnowball are first trained on the supervised seed set $\mathcal{S}$. Considering $\mathcal{S}$ cannot cover sufficient data to train all modules well, QASnowball thus iteratively generates data. At the $i$-th iteration, the newly formed data set $\mathcal{S}_i$ is used to expand the seed set and fine-tune the modules of QASnowball, and QASnowball goes over the whole process again with the fine-tuned modules in subsequent iterations. Next, we will briefly introduce the three modules of QASnowball and describe how to make the whole framework iteratively self-enhanced.

\subsection{Answer Extractor}

Given a document $D$, we treat extracting the core phrases in $D$ that are most likely to be answers as a sequence labeling problem.
For each token in $D$, we ask the extractor to determine whether the token is part of an answer candidate, and merge adjacent tokens predicted to be part of an answer to form an answer candidate.
Specifically, we adopt BERT~\cite{devlin2019bert} as the backbone model of our extractor, which is a typical auto-encoding pre-trained language model (PLM) built on the encoder architecture, and then build a binary classifier based on the backbone to identify whether each token is part of an answer. In order to facilitate the subsequent introduction of our framework, we formalize the extraction process as
\begin{equation*}
\mathcal{A}_D = \{A_D^1, A_D^2, \ldots, A_D^n\} = \texttt{AE}(D).
\end{equation*}
The extractor is first trained on the supervised seed samples, taking the documents and answers of these seed samples as supervision signals. During iterative data generation, the extractor can be improved with newly generated data.

\subsection{Question Generator}

Given a document $D$ and an answer candidate $A_D^i$ extracted from $D$, we treat generating the question $Q_D^i$ corresponding to $A_D^i$ as a contextual generation problem.
Specifically, we select T5~\cite{raffel2020exploring} as the generator backbone, which is a typical generative PLM built on the encoder-decoder architecture. 
Before generating the question, we first add special tags before and after the span of $A_D^i$ to highlight the position of the answer candidate. 
For example, given the document \textit{``... Mark Twain was born on November 30, 1835, in Florida, Missouri ...''} and the answer candidate \textit{``Florida, Missouri''}, we add two special tags (i.e., \textit{``<ANS>''} and \textit{``</ANS>''}) and then input \textit{``... Mark Twain was born on November 30, 1835, in <ANS> Florida, Missouri </ANS> ...''} into the generator. 
By feeding the tagged document into the encoder of the generator, we require the decoder of the generator to output the corresponding question. We formalize the generation process as
\begin{equation*}
\mathcal{Q}_D = \{Q_D^1, Q_D^2, \ldots, Q_D^n\} = \texttt{QG}(D, \mathcal{A}_D).
\end{equation*}
Similar to the answer extractor, the question generator is first trained on the seed set, by taking the documents and answers of seed samples as input and questions as output supervision signals. The newly generated data will also be used to fine-tune the generator, enabling the generator to generate more diverse questions beyond the domain constraints of the initial seed set.

\subsection{Data Filter}

Generally, existing methods usually train a QA model to filter the generated data~\cite{lewis2021paq}. Given a document $D$ and its candidate question-answer pair $(Q_D^i, A_D^i)$, these filtering methods adopt QA models to directly give the reasonableness score of $A_D^i$ based on the semantics of $D$ and $Q_D^i$. Based on the reasonableness scores, these filtering methods can select high-quality data. 
Although these methods have achieved promising results, the filtering quality of these methods depends heavily on the performance of the QA models, and it is difficult to select out those samples that are correctly generated but judged ambiguously by the QA models, which obviously contradicts the purpose of generating QA data to further enhance the QA model.

In this paper, we build the data filter based on the ensemble of the QA model and heuristic rules. Specifically, we choose the recent competitive QA model PERT~\cite{cui2022pert} as the backbone of the filter. Instead of giving the reasonableness score of $A_D^i$, we require the filter backbone to generate an answer $\tilde{A}_D^i$ based on $D$ and $Q_D^i$, without considering $A_D^i$. Then, several heuristic rules are adopted to select and modify the data:
\begin{itemize}
\item if $A_D^i$ does not overlap with $\tilde{A}_D^i$ any more, we discard this example.
\item if $A_D^i$ matches $\tilde{A}_D^i$ exactly, the example $\big(D, Q_D^i, A_D^i\big)$ is finally generated.
\item If $A_D^i$ partially overlaps with $\tilde{A}_D^i$, the example of $\big(D, Q_D^i, [A_D^i; \tilde{A}_D^i]\big)$ is finally generated, where $[A_D^i; \tilde{A}_D^i]$ denotes the combination of $A$ and $\tilde{D}$, such as combining \textit{``Florida,''} and \textit{``Missouri''} into \textit{``Florida, Missouri''}.
\end{itemize}
For convenience, we formalize the filtering as
\begin{equation*}
\mathcal{G}_D = \texttt{F}(D, \mathcal{A}_D, \mathcal{Q}_D),
\end{equation*}
where $\mathcal{G}_D$ is the final generated example set. Compared with the existing filtering methods based entirely on QA models, our filter not only ensures the data quality, but also retains some generated data inconsistent with QA models, which can bring better data diversity. 

\begin{algorithm}[t]
    \caption{The iterative generation process of QASnowball}
    \label{alg:AOA}
    \renewcommand{\algorithmicrequire}{\textbf{Input:}}
    \renewcommand{\algorithmicensure}{\textbf{Output:}}
    \begin{algorithmic}[1]
        \REQUIRE The seed set $\mathcal{S}$, the unlabeled document set $\mathcal{D}$, and the iteration number $t$
        \STATE  Divide the unlabeled document set into several parts according to $t$, and obtain $\mathcal{D}_{1}, \mathcal{D}_{2}, \ldots, \mathcal{D}_{t}$
        \FOR{$i \in [1, t]$}
            \IF{$i = 1$}
                \STATE Train the extractor $\texttt{AE}$, the generator $\texttt{QG}$, and the filter $\texttt{F}$ with $\mathcal{S}$
            \ELSE
                \STATE Fine-tune the extractor $\texttt{AE}$, the generator $\texttt{QG}$, and the filter $\texttt{F}$ with $\mathcal{S}$
            \ENDIF
            \STATE $\mathcal{S}_i \leftarrow \emptyset$
            \FOR{$D \in \mathcal{D}_{i}$}
                \STATE $\mathcal{A}_D \leftarrow  \texttt{AE}(D)$
                \STATE $\mathcal{Q}_D \leftarrow  \texttt{QG}(D, \mathcal{A}_D)$
                \STATE $\mathcal{G}_D \leftarrow  \texttt{F}(D, \mathcal{A}_D, \mathcal{Q}_D)$
                \STATE $\mathcal{S}_i \leftarrow \mathcal{S}_i \cup \mathcal{G}_D$
            \ENDFOR
            \STATE Update $\mathcal{S}$ with $\mathcal{S}_i$
        \ENDFOR
        \ENSURE The generated data of all iterations $\mathcal{S}_1$, $\mathcal{S}_2$, $\ldots$, $\mathcal{S}_t$, the answer extractor $\texttt{AE}$, the question generator $\texttt{QG}$, the data filter $\texttt{F}$.
    \end{algorithmic}
\end{algorithm}

\subsection{Iterative Bootstrapping Process}

As we mentioned before, using only the initial supervised seed set to build the generation system and generate data at one time may suffer from diversity and coverage issues. Therefore, QASnowball takes an iterative approach to generate data and dynamically fine-tunes the QASnowball framework during the generation process.
Specifically, given an unlabeled document set $\mathcal{D}$ that requires QASnowball to annotate questions and answers, QASnowball first divides the document set into several parts $\mathcal{D}_1, \mathcal{D}_2, \ldots, \mathcal{D}_t$ according to the total number of iterations $t$. At the $i$-th iteration, QASnowball automatically annotates the questions and answers for $\mathcal{D}_i$ and generates the dataset $\mathcal{S}_i$, the whole process is given as
\begin{equation*}
\mathcal{S}_i = \cup_{D \in \mathcal{D}_i} \mathcal{G}_D.
\end{equation*}
After obtaining $\mathcal{S}_i$, we update the seed set $\mathcal{S}$ with $\mathcal{S}_i$, and the updated seed set is used to fine-tune the modules of QASnowball. In general, the updating strategy can be to replace $\mathcal{S}$ with $\mathcal{S}_i$ or to merge $\mathcal{S}$ and $\mathcal{S}_i$. Due to our pilot experiments, we merge $\mathcal{S}$ and $\mathcal{S}_i$ to update the seed set in this paper. In fact, more sophisticated strategies can be used here to update the seed set as well, and we leave this for future work. Algorithm~\ref{alg:AOA} shows the iterative bootstrapping process of QASnowball.

\section{Experiments}

To evaluate the effectiveness of QASnowball and the quality of the data generated by QASnowball, we conduct experiments in both the high-resource English and the medium-resource Chinese scenarios. For convenience, we name the data generated by QASnowball with ``\textbf{S}nowball \textbf{G}enerated \textbf{Q}A data (SGQ)''.

\subsection{Datasets}

For the English scenario, we conduct experiments on 
SQuAD~\cite{rajpurkar2016squad}, TriviaQA-wiki~\cite{joshi2017triviaqa}, TriviaQA-web~\cite{joshi2017triviaqa}, and HotpotQA~\cite{yang2018hotpotqa}.
Among these datasets, we select SQuAD, TriviaQA-wiki, and TriviaQA-web as the seed set to automatically generate large-scale English QA data and remain HotpotQA to evaluate the transferability of the generated English data, i.e., to evaluate whether our framework can generate some data out of the distribution of the initial seed set.
Besides the above datasets, we also compare our SGQ with PAQ~\cite{lewis2021paq}, a recently proposed auto-labeled QA dataset.

For the Chinese scenario, we conduct experiments on CMRC~\cite{cui2019span}, DuReader~\cite{he2018dureader}, SQuAD-zh~\footnote{\url{https://github.com/junzeng-pluto/ChineseSquad/tree/master/squad-zen}}, and DRCD\cite{shao2018drcd}. 
Among these datasets, we select CMRC, DuReader, and SQuAD-zh as the seed set to automatically generate large-scale Chinese QA data and remain DRCD to evaluate the transferability of the generated Chinese data.

For SGQ, we use QASnowball to automatically generate 1 million Chinese samples and 1 million English samples through two rounds of iterations.

\begin{figure}[t]
  \subfigure[The EM scores of PERT trained on PAQ, SGQ and supervised English datasets.]{
  \begin{minipage}{0.45\linewidth}
    \includegraphics[width=\linewidth]{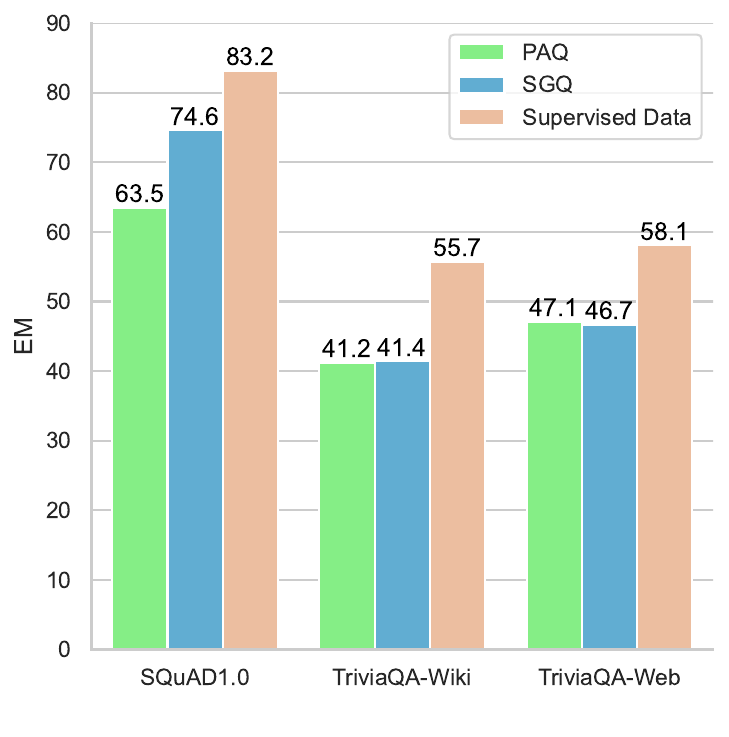}
  \end{minipage}}
   \hspace{0.02\linewidth}
  \subfigure[The F1 scores of PERT trained on PAQ, SGQ and supervised English datasets.]{
  \begin{minipage}{0.45\linewidth}
    \includegraphics[width=\linewidth]{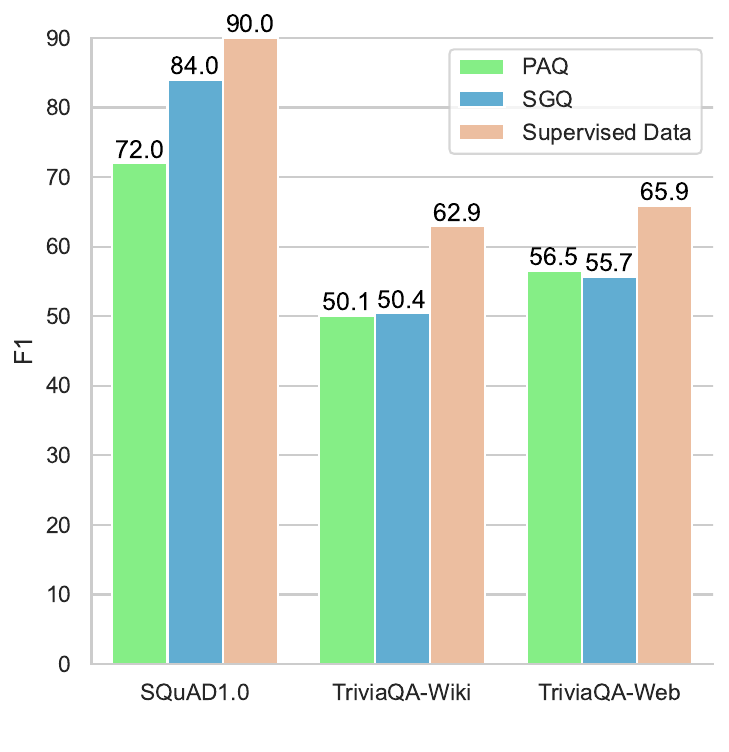}
  \end{minipage}}
  \caption{The performance comparison of training PERT on SGQ and other English QA datasets.}
  \label{fig:zero-supervised-1}
\end{figure}

\begin{figure}[t]
  \subfigure[The EM scores of PERT trained on SGQ and supervised Chinese datasets.]{
  \begin{minipage}{0.43\linewidth}
    \includegraphics[width=\linewidth]{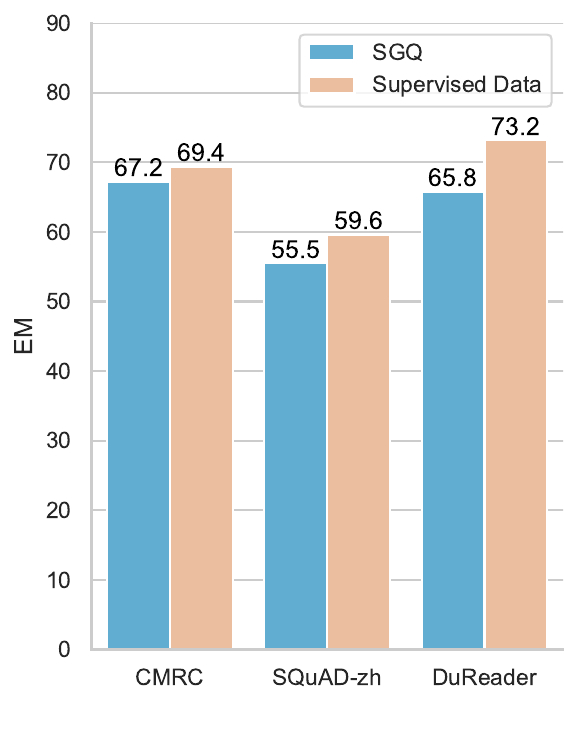}
  \end{minipage}}
  \hspace{0.05\linewidth}
  \subfigure[The F1 scores of PERT trained on SGQ and supervised Chinese  datasets.]{
  \begin{minipage}{0.45\linewidth}
    \includegraphics[width=\linewidth]{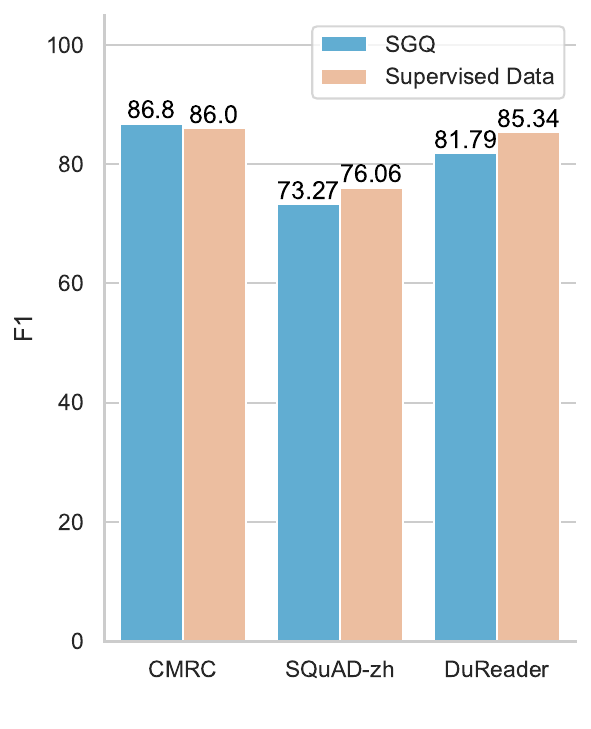}
  \end{minipage}}
  \caption{The performance comparison of training PERT on SGQ and other Chinese QA datasets.}
  \label{fig:zero-supervised-2}
\end{figure}

\begin{table}[t]
\centering
\scalebox{0.95}{
\begin{tabular}{lllcc}
\toprule
Dataset                   & Shot                & Metric & Seed & Seed+SGQ \\
\midrule
\multirow{6}{*}{HotpotQA} & \multirow{2}{*}{0}  & EM     & 53.1 & \textbf{55.3}   \\
                          &                     & F1     & 66.6 & \textbf{68.0 }   \\
\cmidrule{2-5}
                          & \multirow{2}{*}{8}  & EM     & 53.1 & \textbf{55.7 }   \\
                          &                     & F1     & 66.6 & \textbf{68.6 }   \\
\cmidrule{2-5}
                          & \multirow{2}{*}{32} & EM     & 55.0 & \textbf{56.5 }   \\
                          &                     & F1     & 68.9 & \textbf{69.6 }   \\
\midrule
\multirow{6}{*}{DRCD}     & \multirow{2}{*}{0}  & EM     & 71.8 & \textbf{74.3}   \\
                          &                     & F1     & 85.1 & \textbf{86.4 }   \\
\cmidrule{2-5}
                          & \multirow{2}{*}{8}  & EM     & 73.3 & \textbf{74.7 }   \\
                          &                     & F1     & 85.9 & \textbf{86.6 }   \\
\cmidrule{2-5}
                          & \multirow{2}{*}{32} & EM     & 74.9 & \textbf{76.3 }   \\
                          &                     & F1     & 86.6 & \textbf{87.4 }   \\
\bottomrule
\end{tabular}}
\caption{\label{tab:hotpot-drcd}
The performance of PERT on HotpotQA and DRCD, without using any training samples from HotpotQA or DRCD. The SGQ here consists of 1 million generated QA samples, which are obtained by 2 rounds of iterations. For HotpotQA, the seed set consists of SQuAD, TriviaQA-wiki, and TriviaQA-web. For DRCD, the seed set consists of CMRC, SQuAD-zh, and DuReader.}
\end{table}

\subsection{Experimental Settings}

Since our main focus is to evaluate the quality of SGQ, we conduct both English and Chinese experiments under the MRC setting. 
On the one hand, this is because MRC is an important basis of QA. On the other hand, the MRC setting is simple and operational, as all answers can be found in documents under this setting. Considering that not all the above datasets are constructed under the MRC setting, we remove all the examples in these datasets whose answers are not included in their documents.
Evaluating the quality of datasets requires an effective QA model as the benchmark model. By training the benchmark model on different datasets, the performance of the benchmark model can reflect the quality of the training data.
In our experiments, we select PERT as our benchmark model because it is a competitive QA model proposed recently.
We use exact match (EM) and F1 scores as the evaluation metrics to evaluate the model performance, both of which are typical metrics adopted in MRC.

\subsection{The Quality of SGQ}

To examine the quality of our SGQ dataset, for each supervised dataset, we train the PERT model on SGQ and the dataset respectively, and then compare the performance difference. For English supervised datasets, we additionally train the PERT model on PAQ. From Figure~\ref{fig:zero-supervised-1} and Figure~\ref{fig:zero-supervised-2}, we find that:

(1) Although training the PERT model only on the automatically generated SGQ does not achieve better performance than training the model on the supervised datasets, these model performances are close, indicating that SGQ is of high quality.

(2) Compared with PAQ, our SGQ can help the PERT model achieve better performance, indicating SGQ is better than PAQ. There are two reasons why SGQ is better than PAQ, one is that the self-enhanced mechanism of QASnowball leads to better data quality, and the other is that the answer extractor of QASnowball can extract long answers rather than named entities in PAQ. Both of these two aspects will be further described in the subsequent sections.

\begin{table*}[t]
\centering
\begin{tabular}{l|cc|cc|cc|cc|cc}
\toprule
\multirow{2}{*}{Dataset} & \multicolumn{2}{c|}{CMRC} & \multicolumn{2}{c|}{SQuAD-zh} & \multicolumn{2}{c|}{DuReader} & \multicolumn{2}{c|}{DRCD} & \multicolumn{2}{c}{Average} \\
                         & EM          & F1         & EM            & F1           & EM            & F1           & EM          & F1         & EM         & F1         \\
\toprule
$\text{SGQ}_1$  & 67.2        & 86.8       & 55.5          & 73.3         & 65.8          & 81.8         & 67.8        & 83.4       & 64.1     & 81.3     \\
$\text{SGQ}_2$  & \textbf{71.6}        & \textbf{88.0}         & \textbf{57.8}          & \textbf{74.8}         & \textbf{70.9}          & \textbf{83.7}         & \textbf{70.5}        & \textbf{84.7}       & \textbf{67.7}       & \textbf{82.8}       \\
$\text{SGQ}_1-\text{SGQ}_2$ & 33.4        & 78.5       & 40.7          & 66.0           & 34.1          & 72.5         & 45.6        & 73.2       & 38.5      & 72.6      \\
$\text{SGQ}_2-\text{SGQ}_1$ & 70.9        & 85.5       & 41.9          & 66.3         & 66.3          & 79.9         & 68.4        & 81.1       & 61.9     & 78.2      \\
\bottomrule
\end{tabular}
\caption{\label{tab:zero-shot}
The performance of PERT models trained on datasets generated by QASnowball with different iterations. $\text{SGQ}_1$ indicating 1 million samples generated through only one iteration, i.e., generating data without iteratively self-enhanced. $\text{SGQ}_2$ indicating 1 million samples generated through two rounds of iterations.
}
\end{table*}

\begin{table*}[t]
\centering
\begin{tabular}{c|cc| cc| cc| cc| cc}
\toprule
\multirow{2}{*}{Size of SGQ} & 
\multicolumn{2}{c|}{{CMRC}} &
\multicolumn{2}{c|}{{SQuAD-zh}} &
\multicolumn{2}{c|}{{DuReader}} &
\multicolumn{2}{c|}{{DRCD}} & 
\multicolumn{2}{c}{Average}\\
& EM & F1 & EM & F1 & EM & F1 & EM & F1 & EM & F1\\ 
\midrule
0.0M & 69.4 & 86.0  & 59.6 & 76.1 & 73.2 & 85.3 & 88.7 & 93.9  & 72.8 & 85.3\\
\midrule
0.1M & \textbf{73.6} & 88.3  & 60.2 & 76.6 & 74.0 & 86.3 & 88.1 & 93.5  & 74.0 & 86.2\\
0.5M & 72.8 & 89.3  & 60.6 & 77.0 & 75.0 & 86.6 & 89.3 & \textbf{94.3}  & 74.5 & 86.8\\
1.0M & 73.3 & 89.5  & 60.3 & 76.7 & \textbf{76.1} & \textbf{86.9} & \textbf{89.6} & \textbf{94.3}  & 74.8 & 86.9\\
5.0M & 73.4 & \textbf{90.1} & \textbf{60.8} & \textbf{77.4} & 75.7 & 86.8 & 89.5  & \textbf{94.3}  & \textbf{74.9} & \textbf{87.2}\\
\bottomrule
\end{tabular}
\caption{\label{data-amount}
The results of pre-training the model on SGQ of different sizes and then fine-tuning the model on the supervised datasets. 0.0M indicates directly training the model on the supervised datasets.
}
\end{table*}

To evaluate the transferability of SGQ, we first train the PERT model on the seed set (SQuAD, TriviaQA-wiki, and TriviaQA-web for English; CMRC, SQuAD-zh, and DuReader for Chinese) and SGQ, and then fine-tune the PERT model on HotpotQA and DRCD under the few-shot setting. From Table~\ref{tab:hotpot-drcd}, we find that:

(1) SGQ has great transferability. Even if the PERT model does not use any training samples of HotpotQA and DRCD, the model relies only on other supervised datasets and SGQ can also handle the test samples of HotpotQA and DRCD.

(2) Moreover, the performance of the PERT model is greatly improved after training on SGQ, indicating that the auto-labeled data in SGQ has a broader domain coverage than the supervised seed set and thus yields a significant performance improvement.

\subsection{The Effect of Self-Enhanced Mechanism}

To evaluate the effectiveness of the iterative self-enhanced mechanism of QAsnowball, we compare the quality of the generated SGQ datasets with and without iteratively fine-tuning the framework of QAsnowball. Specifically, we first use only one iteration to complete the generation of all 1 million samples, i.e., the entire data generation process does not involve updating QAsnowball. We denote the dataset generated with one iteration as $\text{SGQ}_1$. Then, we use two iterations to generate 1 million samples based on the same unlabeled corpus. Specifically, $\text{SGQ}_1$ is further used to fine-tune QASnowball for generating the final dataset, denoted as $\text{SGQ}_2$. We train the PERT model on $\text{SGQ}_1$ and $\text{SGQ}_2$ respectively, and then compare the performance difference. As shown in Table~\ref{tab:zero-shot}, we can clearly find that:

(1) The performance of PERT models trained on SGQ$_2$ is consistently better than ones trained on SQG$_1$. This demonstrate that the quality of the data generated by the iterative mechanism is significantly higher than the data generated without any iterations, indicating the influence of the iterative self-enhanced mechanism on improving the data quality.

(2) We also compare the performance of models trained on $\text{SGQ}_1-\text{SGQ}_2$ and $\text{SGQ}_2-\text{SGQ}_1$. The results show that the quality of the samples that are in $\text{SGQ}_2$ but not in $\text{SGQ}_1$ is much higher than those samples only in $\text{SGQ}_1$. This implies that there are certain incorrectly generated samples in $\text{SGQ}_1$, and these noise samples can significantly weaken the model performance. Without using the iterative self-enhanced mechanism, we cannot get rid of these noise samples.

\subsection{The Effect of Data Size}

To further show the impact of data size on the performance of QA models, we generate samples more than the above mentioned 1 million samples, and use the pre-training and fine-tuning paradigm with both the auto-labeled and supervised data. Specifically, we first pre-train the PERT model on the auto-labeled dataset SGQ, and then fine-tune the pre-trained PERT on supervised datasets. All results are shown in the Table~\ref{data-amount}, from which we can find that: 

(1) Pre-training on the large-scale auto-labeled datasets and then fine-tuning on the supervised datasets can be more effective than training the model directly on the supervised datasets. 

(2) As the generated data size increases, the performance of the QA model increases accordingly. This implies that more auto-labeled data can provide more information beyond the restricted domain of the supervised datasets, which can well supplement the supervised datasets.

\begin{table*}[t]
\small
\centering
\begin{tabular}{p{11cm}|p{3cm}}
\toprule
Passage & Question \\
\midrule
\begin{CJK}{UTF8}{gbsn}探讨电视胸腔镜辅助小切口手术(vamt)在自发性气胸治疗中的应用价值。\colorbox{green}{方法是}\colorbox{yellow}{于患侧腋中线第6或第7肋间置入胸腔镜套管} \colorbox{green}{，在靠近病变位置做长约5~8cm切口经肋间进胸(多在腋前线或腋中线第4~5肋)，} \colorbox{green}{使用普通手术器械与胸腔镜手术器械，在胸腔镜和直视下进行操作。}\end{CJK}  

\textbf{Translation}: This passage discusses the value of using video-assisted thoracoscopic surgery (VAMT) assisted small incision surgery in the treatment of spontaneous pneumothorax. \colorbox{green}{The method involves} \colorbox{yellow}{placing a thoracoscope sheath in the midline of the 6th or 7th intercostal space on the} \colorbox{yellow}{ affected side} \colorbox{green}{, making a 5-8cm incision near the affected area through the intercostal} \colorbox{green}{space (usually on the anterior or midline of the 4th or 5th intercostal space) and using} \colorbox{green}{both conventional surgical instruments and thoracoscopic instruments to operate under} \colorbox{green}{thoracoscope and direct vision.} & 
\begin{CJK}{UTF8}{gbsn}
电视胸腔镜辅助小切口手术的方法是什么?
\end{CJK}

\textbf{Translation}:What is the method of video-assisted thoracoscopic small incision surgery?
\\
\hline

\begin{CJK}{UTF8}{gbsn}移动学习逐渐成为一种趋势，而\colorbox{yellow}{移动学习资源开发}\colorbox{red}{是开展移动学习至关重要}的环节。文章详细分析了目前移动学习资源开发中流行的几种移动开发平台及开发工具，并针对其是否适合移动学习资源开发而进行了评价总结。\end{CJK}

\textbf{Translation}: Mobile learning is becoming a trend, and \colorbox{yellow}{the development of mobile learning resources} \colorbox{red}{is a crucial aspect of} conducting mobile learning. The article analyzes several popular mobile development platforms and tools currently used in mobile learning resource development and evaluates their suitability for mobile learning resource development. & 
\begin{CJK}{UTF8}{gbsn}
开展移动学习有哪些环节?
\end{CJK}

\textbf{Translation}: What is the aspect of mobile learning?
\\
\bottomrule
\end{tabular}
\caption{\label{tab:iteration-comparison}
The examples of the data generated in SGQ$_1$ and SGQ$_2$. The answer spans extracted in both SGQ$_1$ and SGQ$_2$ are colored yellow. The additional spans extracted in SGQ$_1$ and SGQ$_2$ are colored red and green, respectively. 
}
\end{table*}

\begin{table*}[t]
\small
\centering
\begin{tabular}{p{12cm}|p{3cm}}
\toprule
Passage & Question \\
\midrule
   \begin{CJK}{UTF8}{gbsn}天然铀矿包含三种同位素：大部分(99.274\%)为铀238，\colorbox{red}{约}\colorbox{yellow}{0.72}\colorbox{green}{\%}的铀235以及约0.0055\%的铀234。如果天然铀被提纯到包含3\%的铀235，那么就可以被轻水反应堆用作燃料。\end{CJK}
   
   \textbf{Translation}: Natural uranium ore contains three isotopes: mostly (99.274\%) uranium 238, \colorbox{red}{about}\colorbox{yellow}{0.72}\colorbox{green}{\%} uranium 235 and about 0.0055\% uranium 234. If natural uranium is purified to contain 3\% uranium-235, it can be used as fuel in light water reactors. & \begin{CJK}{UTF8}{gbsn}天然铀矿含有多少百分比的铀235？\end{CJK} 
   
   \textbf{Translation}: What percentage of uranium 235 does natural uranium ore contain? \\
\bottomrule
\end{tabular}
\caption{\label{tab:filter-comparison}
The examples of the original answers before filtering, the filter's answer, and the new answers after filtering. The overlapping texts between the original answer and the filter's answer are colored yellow. Texts that only appear in the original answer and the filter's answer are colored red and green, respectively.
}
\end{table*}

\subsection{Human Evaluation}

To conduct human evaluation, we randomly select 200 sample pairs from $\text{SGQ}_1$ and $\text{SGQ}_2$, and each pair of them have the same document and question but different answers. 
For the samples selected from $\text{SGQ}_1$, 73\% of these samples have reasonable answers, i.e., 73\% of these generated samples meet human standards.
For the samples selected from $\text{SGQ}_2$, 82\% of these samples have reasonable answers. This demonstrates the effectiveness of QAsnowball which can iteratively improve the quality of generated data.

We further manually analyze the sample pairs that have reasonable answers and observe that 75.3\% of the samples in $\text{SGQ}_2$ have better answers than those in $\text{SGQ}_1$, showing the effectiveness of the self-enhanced mechanism.

\subsection{Case Study}
We show two examples of the data generated in SGQ$_1$ and SGQ$_2$ in Table~\ref{tab:iteration-comparison}. For the first example, we can see that the answer extracted in SGQ$_2$ is more complete than the one in SGQ$_1$. For the second example, it can be seen that the answer extracted in SGQ$_2$ is more concise than the one in SGQ$_1$. These indicate that the iterative self-enhanced mechanism of QASnowball can improve the quality of the generated data. Table~\ref{tab:filter-comparison} shows that using the ensemble of the QA model and rules to build the data filter can lead to better generation.



\section{Conclusion}

In this paper, we propose QASnowball, a novel iterative bootstrapping framework that can continually generate high-quality and large-scale QA data. To evaluate the effectiveness of QASnowball and its generated data, we conduct sufficient experiments in both the high-resource English and the medium-resource Chinese scenarios. The experimental results show that the dataset SGQ generated by QASnowball can be comparable to some supervised datasets. Further experiments on the Chinese datasets demonstrate that pre-training the model on SGQ and then fine-tuning the pre-trained model on supervised datasets can have better results than only using supervised datasets. As compared with the existing auto-labeled dataset PAQ, which is built with a non-iterative method, the model trained on SGQ can also outperform the model trained on PAQ. Moreover, our empirical analysis and case study demonstrate the necessity and importance of the self-enhanced mechanism for generating high-quality data. 

In the future, we will release both the framework QASnowball and the dataset SGQ, and further explore the following directions:

(1) Although we demonstrate the feasibility of iteratively generating QA data, many internal settings, such as how to update the seed set, still need further exploration. 

(2) The recent works for chain-of-thought prompting~\cite{wei2022chain} have drawn much attention from researchers to a more complex QA scenario that requires reasoning. How to acquire more complex QA data to build a reasoning system is worthy of further study.

\section*{Limitations}

A limitation of this work is that this work mainly focuses on verifying the feasibility of iteratively generating QA data, and the exploration of the internal mechanisms that make the iterative generation effective still needs to be further advanced. Exploring the limits of data quality improvement as the number of iterations increases and showing how the iterative process actually helps each part of the model may have important implications for the practical use of our framework. In addition to further exploring the internal mechanisms of our iterative bootstrapping framework, the efficiency of the question generator and the data filter also needs to be further improved, and the efficiency issues of these modules largely limit the size of the generated data. For these limitations, we will do our best to address them in the future.

\section*{Ethics Statement}

The authors of this work declare that they have no conflict of interest. Besides, no animal or human being is involved as the study objective in any part of this article.

\bibliography{acl2023}

\begin{thebibliography}{26}
\expandafter\ifx\csname natexlab\endcsname\relax\def\natexlab#1{#1}\fi

\bibitem[{Agichtein and Gravano(2000)}]{agichtein2000snowball}
Eugene Agichtein and Luis Gravano. 2000.
\newblock Snowball: Extracting relations from large plain-text collections.
\newblock In \emph{Proceedings of JCDL}, pages 85--94.

\bibitem[{Alberti et~al.(2019)Alberti, Andor, Pitler, Devlin, and
  Collins}]{alberti2019synthetic}
Chris Alberti, Daniel Andor, Emily Pitler, Jacob Devlin, and Michael Collins.
  2019.
\newblock Synthetic {QA} corpora generation with roundtrip consistency.
\newblock In \emph{Proceedings of ACL}, pages 6168--6173.

\bibitem[{Baradaran et~al.(2022)Baradaran, Ghiasi, and
  Amirkhani}]{baradaran2022survey}
Razieh Baradaran, Razieh Ghiasi, and Hossein Amirkhani. 2022.
\newblock A survey on machine reading comprehension systems.
\newblock \emph{Natural Language Engineering}, 28(6):683--732.

\bibitem[{Brown et~al.(2020)Brown, Mann, Ryder, Subbiah, Kaplan, Dhariwal,
  Neelakantan, Shyam, Sastry, Askell et~al.}]{brown2020language}
Tom Brown, Benjamin Mann, Nick Ryder, Melanie Subbiah, Jared~D Kaplan, Prafulla
  Dhariwal, Arvind Neelakantan, Pranav Shyam, Girish Sastry, Amanda Askell,
  et~al. 2020.
\newblock Language models are few-shot learners.
\newblock In \emph{Proceedings of NeurIPS}, pages 1877--1901.

\bibitem[{Chen et~al.(2017)Chen, Fisch, Weston, and Bordes}]{chen2017reading}
Danqi Chen, Adam Fisch, Jason Weston, and Antoine Bordes. 2017.
\newblock Reading wikipedia to answer open-domain questions.
\newblock In \emph{Proceedings of ACL}, pages 1870--1879.

\bibitem[{Cui et~al.(2019)Cui, Liu, Che, Xiao, Chen, Ma, Wang, and
  Hu}]{cui2019span}
Yiming Cui, Ting Liu, Wanxiang Che, Li~Xiao, Zhipeng Chen, Wentao Ma, Shijin
  Wang, and Guoping Hu. 2019.
\newblock A span-extraction dataset for chinese machine reading comprehension.
\newblock In \emph{Proceedings of EMNLP-IJCNLP}, pages 5883--5889.

\bibitem[{Cui et~al.(2022)Cui, Yang, and Liu}]{cui2022pert}
Yiming Cui, Ziqing Yang, and Ting Liu. 2022.
\newblock {PERT}: Pre-training bert with permuted language model.
\newblock \emph{arXiv preprint arXiv:2203.06906}.

\bibitem[{Devlin et~al.(2019)Devlin, Chang, Lee, and
  Toutanova}]{devlin2019bert}
Jacob Devlin, Ming{-}Wei Chang, Kenton Lee, and Kristina Toutanova. 2019.
\newblock {BERT:} pre-training of deep bidirectional transformers for language
  understanding.
\newblock In \emph{Proceedings of NAACL-HLT}, pages 4171--4186.

\bibitem[{Fang et~al.(2020)Fang, Wang, Gan, Sun, Liu, and
  Zhu}]{fang2020accelerating}
Yuwei Fang, Shuohang Wang, Zhe Gan, Siqi Sun, Jingjing Liu, and Chenguang Zhu.
  2020.
\newblock Accelerating real-time question answering via question generation.
\newblock \emph{arXiv preprint arXiv:2009.05167}.

\bibitem[{Gao et~al.(2020)Gao, Han, Xie, Liu, Lin, Lin, and
  Sun}]{gao2020neural}
Tianyu Gao, Xu~Han, Ruobing Xie, Zhiyuan Liu, Fen Lin, Leyu Lin, and Maosong
  Sun. 2020.
\newblock Neural snowball for few-shot relation learning.
\newblock In \emph{Proceedings of AAAI}, pages 7772--7779.

\bibitem[{He et~al.(2018)He, Liu, Liu, Lyu, Zhao, Xiao, Liu, Wang, Wu, She
  et~al.}]{he2018dureader}
Wei He, Kai Liu, Jing Liu, Yajuan Lyu, Shiqi Zhao, Xinyan Xiao, Yuan Liu,
  Yizhong Wang, Hua Wu, Qiaoqiao She, et~al. 2018.
\newblock {DuReader}: a {C}hinese machine reading comprehension dataset from
  real-world applications.
\newblock In \emph{Proceedings of the Workshop on Machine Reading for Question
  Answering}, pages 37--46.

\bibitem[{Joshi et~al.(2017)Joshi, Choi, Weld, and
  Zettlemoyer}]{joshi2017triviaqa}
Mandar Joshi, Eunsol Choi, Daniel~S Weld, and Luke Zettlemoyer. 2017.
\newblock {TriviaQA}: A large scale distantly supervised challenge dataset for
  reading comprehension.
\newblock In \emph{Proceedings of ACL}, pages 1601--1611.

\bibitem[{Karpukhin et~al.(2020)Karpukhin, Oguz, Min, Lewis, Wu, Edunov, Chen,
  and Yih}]{karpukhin2020dense}
Vladimir Karpukhin, Barlas Oguz, Sewon Min, Patrick Lewis, Ledell Wu, Sergey
  Edunov, Danqi Chen, and Wen-tau Yih. 2020.
\newblock Dense passage retrieval for open-domain question answering.
\newblock In \emph{Proceedings of EMNLP}, pages 6769--6781.

\bibitem[{Lewis et~al.(2021)Lewis, Wu, Liu, Minervini, K{\"u}ttler, Piktus,
  Stenetorp, and Riedel}]{lewis2021paq}
Patrick Lewis, Yuxiang Wu, Linqing Liu, Pasquale Minervini, Heinrich
  K{\"u}ttler, Aleksandra Piktus, Pontus Stenetorp, and Sebastian Riedel. 2021.
\newblock {PAQ}: 65 million probably-asked questions and what you can do with
  them.
\newblock \emph{Transactions of the Association for Computational Linguistics},
  9:1098--1115.

\bibitem[{Li et~al.(2020)Li, Feng, Meng, Han, Wu, and Li}]{li2020unified}
Xiaoya Li, Jingrong Feng, Yuxian Meng, Qinghong Han, Fei Wu, and Jiwei Li.
  2020.
\newblock A unified mrc framework for named entity recognition.
\newblock In \emph{Proceedings of ACL}, pages 5849--5859.

\bibitem[{Liu et~al.(2020)Liu, Chen, Liu, Bi, and Liu}]{liu2020event}
Jian Liu, Yubo Chen, Kang Liu, Wei Bi, and Xiaojiang Liu. 2020.
\newblock Event extraction as machine reading comprehension.
\newblock In \emph{Proceedings of EMNLP}, pages 1641--1651.

\bibitem[{Liu et~al.(2019)Liu, Zhang, Zhang, Wang, and Zhang}]{liu2019neural}
Shanshan Liu, Xin Zhang, Sheng Zhang, Hui Wang, and Weiming Zhang. 2019.
\newblock Neural machine reading comprehension: Methods and trends.
\newblock \emph{Applied Sciences}, 9(18):3698.

\bibitem[{Nogueira et~al.(2019)Nogueira, Yang, Lin, and
  Cho}]{nogueira2019document}
Rodrigo Nogueira, Wei Yang, Jimmy Lin, and Kyunghyun Cho. 2019.
\newblock Document expansion by query prediction.
\newblock \emph{arXiv preprint arXiv:1904.08375}.

\bibitem[{Raffel et~al.(2020)Raffel, Shazeer, Roberts, Lee, Narang, Matena,
  Zhou, Li, and Liu}]{raffel2020exploring}
Colin Raffel, Noam Shazeer, Adam Roberts, Katherine Lee, Sharan Narang, Michael
  Matena, Yanqi Zhou, Wei Li, and Peter~J Liu. 2020.
\newblock Exploring the limits of transfer learning with a unified text-to-text
  transformer.
\newblock \emph{Journal of Machine Learning Research}, 21:1--67.

\bibitem[{Rajpurkar et~al.(2016)Rajpurkar, Zhang, Lopyrev, and
  Liang}]{rajpurkar2016squad}
Pranav Rajpurkar, Jian Zhang, Konstantin Lopyrev, and Percy Liang. 2016.
\newblock {SQuAD}: 100,000+ questions for machine comprehension of text.
\newblock In \emph{Proceedings of EMNLP}, pages 2383--2392.

\bibitem[{Shao et~al.(2018)Shao, Liu, Lai, Tseng, and Tsai}]{shao2018drcd}
Chih~Chieh Shao, Trois Liu, Yuting Lai, Yiying Tseng, and Sam Tsai. 2018.
\newblock {DRCD}: a chinese machine reading comprehension dataset.
\newblock \emph{arXiv preprint arXiv:1806.00920}.

\bibitem[{Thoppilan et~al.(2022)Thoppilan, De~Freitas, Hall, Shazeer,
  Kulshreshtha, Cheng, Jin, Bos, Baker, Du et~al.}]{thoppilan2022lamda}
Romal Thoppilan, Daniel De~Freitas, Jamie Hall, Noam Shazeer, Apoorv
  Kulshreshtha, Heng-Tze Cheng, Alicia Jin, Taylor Bos, Leslie Baker, Yu~Du,
  et~al. 2022.
\newblock {LaMDA}: Language models for dialog applications.
\newblock \emph{arXiv preprint arXiv:2201.08239}.

\bibitem[{Wei et~al.(2022)Wei, Wang, Schuurmans, Bosma, Chi, Le, and
  Zhou}]{wei2022chain}
Jason Wei, Xuezhi Wang, Dale Schuurmans, Maarten Bosma, Ed~Chi, Quoc Le, and
  Denny Zhou. 2022.
\newblock Chain of thought prompting elicits reasoning in large language
  models.
\newblock \emph{arXiv preprint arXiv:2201.11903}.

\bibitem[{Yang et~al.(2018)Yang, Qi, Zhang, Bengio, Cohen, Salakhutdinov, and
  Manning}]{yang2018hotpotqa}
Zhilin Yang, Peng Qi, Saizheng Zhang, Yoshua Bengio, William Cohen, Ruslan
  Salakhutdinov, and Christopher~D Manning. 2018.
\newblock {HotpotQA}: A dataset for diverse, explainable multi-hop question
  answering.
\newblock In \emph{Proceedings of EMNLP}, pages 2369--2380.

\bibitem[{Zhu et~al.(2021)Zhu, Lei, Wang, Zheng, Poria, and
  Chua}]{zhu2021retrieving}
Fengbin Zhu, Wenqiang Lei, Chao Wang, Jianming Zheng, Soujanya Poria, and
  Tat-Seng Chua. 2021.
\newblock Retrieving and reading: A comprehensive survey on open-domain
  question answering.
\newblock \emph{arXiv preprint arXiv:2101.00774}.

\bibitem[{Zhu et~al.(2009)Zhu, Nie, Liu, Zhang, and Wen}]{zhu2009statsnowball}
Jun Zhu, Zaiqing Nie, Xiaojiang Liu, Bo~Zhang, and Ji-Rong Wen. 2009.
\newblock Statsnowball: a statistical approach to extracting entity
  relationships.
\newblock In \emph{Proceedings of the 18th international conference on World
  wide web}, pages 101--110.

\end{thebibliography}
\bibliographystyle{acl_natbib}

\clearpage

\end{document}